\documentclass[10pt,twocolumn,letterpaper]{article}

\usepackage{cvpr}

\usepackage{graphicx}
\usepackage{amsmath}
\usepackage{amssymb}
\usepackage{booktabs}

\usepackage[pagebackref,breaklinks,colorlinks]{hyperref}

\usepackage[capitalize]{cleveref}
\crefname{section}{Sec.}{Secs.}
\Crefname{section}{Section}{Sections}
\Crefname{table}{Table}{Tables}
\crefname{table}{Tab.}{Tabs.}

\def\cvprPaperID{7411}
\def\confName{CVPR}
\def\confYear{2023}

\begin{document}

\title{Inversion-Based Style Transfer with Diffusion Models}


\author{Yuxin Zhang$^{1,2}$\hspace{0.2in} Nisha Huang$^{1,2}$\hspace{0.2in} Fan Tang$^{3}$\hspace{0.2in}  Haibin Huang$^4$\\ Chongyang Ma$^4$\hspace{0.2in} Weiming Dong$^{1,2*}$\hspace{0.2in} Changsheng Xu$^{1,2}$\\
$^1$MAIS, Institute of Automation, Chinese Academy of Sciences \hspace{0.2in} $^2$School of AI, UCAS\\
$^3$Institute of Computing Technology, Chinese Academy of Sciences \hspace{0.15in} $^4$Kuaishou Technology \hspace{0.15in} 
}

\twocolumn[{%
\renewcommand\twocolumn[1][]{#1}%
\maketitle
\begin{center}
    \captionsetup{type=figure}
    \vskip -9mm
    \includegraphics[width=1.0\linewidth]{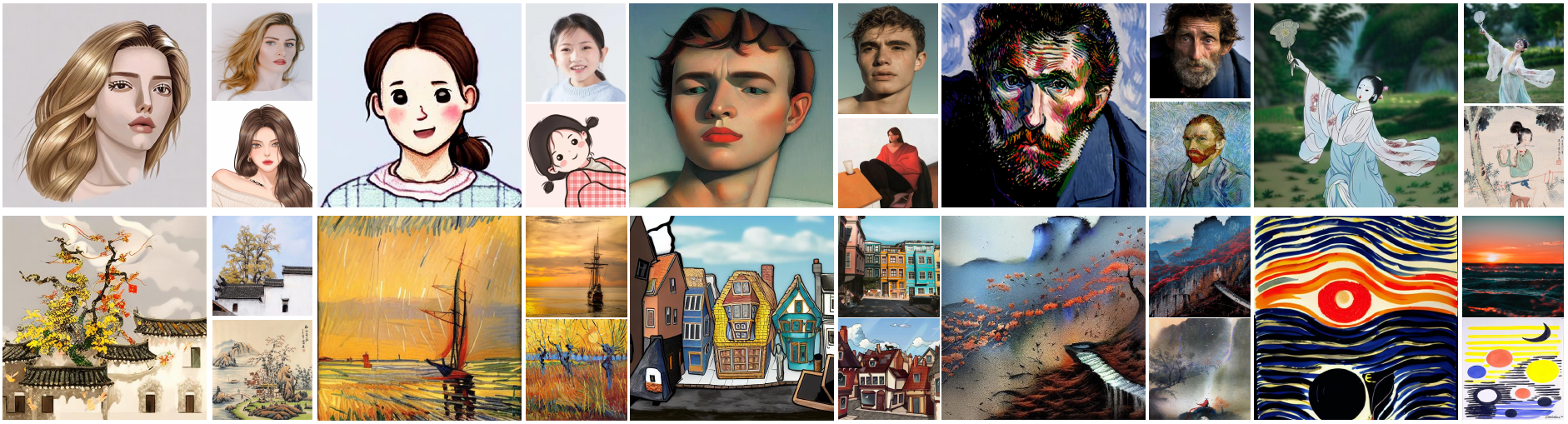}
    \captionof{figure}{
    Style transfer results by using our method.
    Given only a single input painting, our method can accurately transfer the style attributes such as semantics, material, object shape, brushstrokes and colors of the references  to a natural image with a very simple learned textual description ``[C]''. }
    \label{fig:teaser}
    \end{center}%
}]

\renewcommand{\thefootnote}{}
\footnotetext{\textsuperscript{*}Corresponding author: weiming.dong@ia.ac.cn}

\begin{abstract}
The artistic style within a painting is the means of expression, which includes not only the painting material, colors, and brushstrokes, but also the high-level attributes including semantic elements, object shapes, etc.  
Previous arbitrary example-guided artistic image generation methods often fail to control shape changes or convey elements.  
The pre-trained text-to-image synthesis diffusion probabilistic models have achieved remarkable quality, but it often requires extensive textual descriptions to accurately portray attributes of a particular painting. 
We believe that the uniqueness of an artwork lies precisely in the fact that it cannot be adequately explained with normal language.
Our key idea is to learn artistic style directly from a single painting and then guide the synthesis without providing complex textual descriptions.  
Specifically, we assume style as a learnable textual description of a painting.  
We propose an inversion-based style transfer method (InST), which can efficiently and accurately learn the key information of an image, thus capturing and transferring the artistic style of a painting.  
We demonstrate the quality and efficiency of our method on numerous paintings of various artists and styles.
Code and models are available at \url{https://github.com/zyxElsa/InST}.

\end{abstract}

\section{Introduction}

If a photo speaks 1000 words, every painting tells a story.
A painting contains the engagement of an artist's own creation.
The artistic style of a painting can be the personalized textures and brushstrokes, the portrayed beautiful moment or some particular semantic elements.
All those artistic factors are difficult to be described by words.
Therefore, when we wish to utilize a favorite painting to create new digital artworks which can imitate the original idea of the artist, the task turns to example-guided artistic image generation.

\begin{figure*}
\centering
\includegraphics[width=0.99\linewidth]{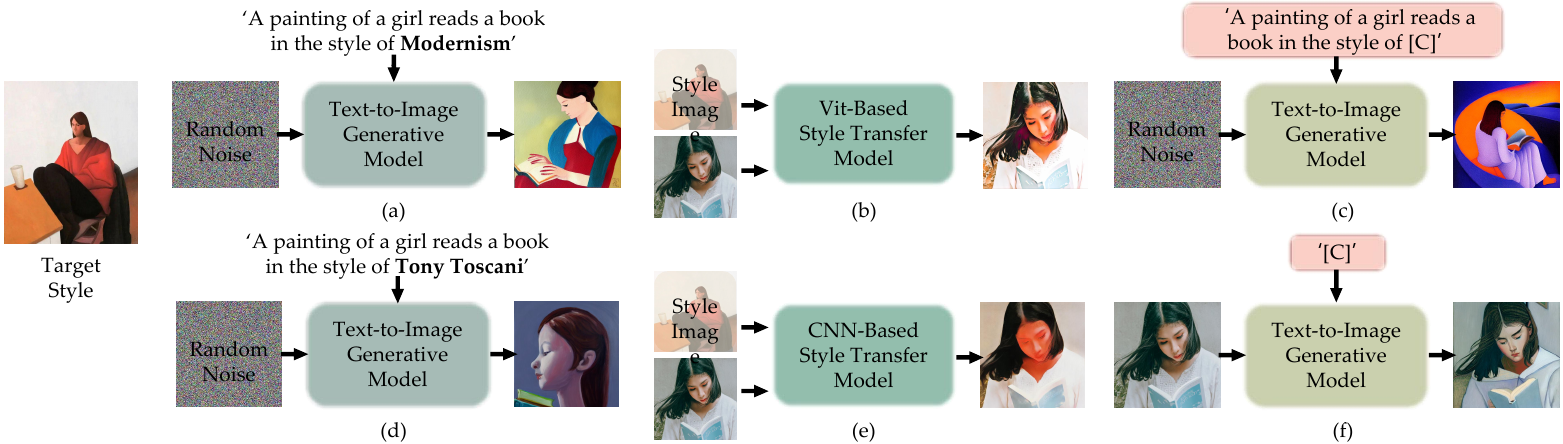}
\caption{
The concept differences between  text-to-image synthesis 
 ~\cite{von-platen-etal-2022-diffusers}, style transfer ~\cite{deng2022stytr2,Zhang:2022:CAST} and our InST.
}
\label{fig:motivation}
\end{figure*}

Generating artistic image(s) from given example(s) has attracted many interests in recent years.
A typical task is style transfer~\cite{gatyscnnstyle,liu2021adaattn,An:2021:Artflow,Wu:2021:SF,Zhang:2022:CAST}, which can create a new artistic image from arbitrary input pair of natural image and painting image, by combining the content of the natural image and the style of the painting image.
However, particular artistic factors such as object shape and semantic elements are difficult to be transferred (see Figures~\ref{fig:motivation}(b) and \ref{fig:motivation}(e)).
Text guided stylization~\cite{CLIPstyler,fu2022ldast,StyleCLIP,StyleGANNADA} produces an artistic image from a natural image and a text prompt, but usually the text prompt for target style can only be a rough description of material (e.g., ``oil'', ``watercolor'', ``sketch''), art movement(e.g. ``Impressionism'',``Cubism''), artist (e.g., ``Vincent van Gogh'', ``Claude Monet'') or a famous artwork (e.g., ``Starry Night'', ``The Scream'').
Diffusion-based methods~\cite{Guided-Diffusion,nichol2022glide,wu2022creative,Huang2022MGAD,imagic} generate high-quality and diverse artistic images based on a text prompt, with or without image examples.
In addition to the input image, a detailed auxiliary textual input is required to guide the generation process if we want to reproduce some vivid contents and styles, which may be still difficult to reproduce the key idea of a specific painting in the result.

In this paper, we propose a novel example-guided artistic image generation framework InST which related to style transfer and text-to-image synthesis, to alleviate all the above problems.
Given only a single input painting image, our method can learn and transfer its style to a natural image with a very simple text prompt (see Figures~\ref{fig:teaser} and \ref{fig:motivation}(f)).
The resulting image exhibit very similar artistic attributes of the original painting, including material, brushstrokes, colors, object shapes and semantic elements, without losing diversity.
Furthermore, we can also control the content of the resulting image by giving a text description (see Figure~\ref{fig:motivation}(c)).

To achieve this merit, we need to obtain the representation of image style, which refers to the set of attributes that appear in the high-level textual description of the image.
We define the textual descriptions as ``new words'' that do not exist in the normal language and get the embeddings via inversion method.
We benefit from the recent success of diffusion models~\cite{latentdiffusion,von-platen-etal-2022-diffusers} and inversion~\cite{chefer2022image,gal2022image}.
We adapt diffusion models in our work as a backbone to be inverted and as a generator in image-to-image and text-to-image synthesis
Specifically, we propose an efficient and accurate textual inversion based on the attention mechanism, which can quickly learn key features from an image, and a stochastic inversion to maintain the semantic of the content image.
We use CLIP~\cite{clip} image embedding to obtain high-quality initial points, and learn key information in the image through multi-layer cross-attention.
Taking an artistic image as a reference, the attention-based inversion module is fed with its CLIP image embedding and then gives its textual embedding.
The diffusion models conditioning on the textual embedding can produce new images with the learned style of the reference.

To demonstrate the effectiveness of InST, we conduct comprehensive experiments, applying our method to numerous images of various artists and styles.
All of the experiments show that our InST produces outstanding results, generating artistic images that both imitate well the style attributes to a high degree, and achieve content consistent with the input natural images or text descriptions.
We demonstrate much improved visual quality and artistic consistency as compared to state-of-the-art approaches.
These outcomes demonstrate the generality, precision and adaptability of our method.
\section{Related Work}

\begin{figure*}
\centering
\includegraphics[width=0.99\linewidth]{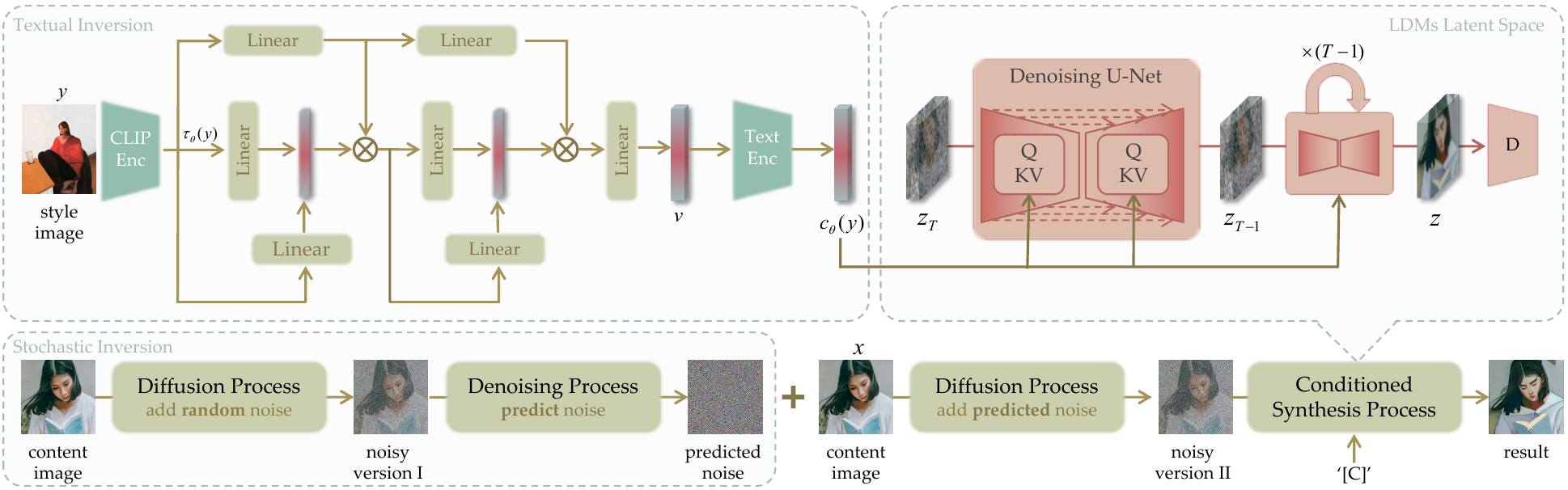}
\caption{
The overview of our InST.
We apply the Stable Diffusion Models~\cite{latentdiffusion,von-platen-etal-2022-diffusers} as the generative backbone and propose an attention-based textual inversion module.
During image synthesis, the inversion module takes the CLIP ~\cite{clip} image embedding $\tau_\theta(y)$ of an artistic image $y$, and gives learned corresponding text embedding $v$, which is then encoded into the standard form of SDMs's caption conditioning $c_\theta(y)$.
Through these, the SDMs can generate new images with encoded latent or random noise $z_T$ conditioned on $c_\theta(y)$.
}
\label{fig:pipeline}
\end{figure*}

\paragraph{Image style transfer}
Image style transfer has been widely studied as a typical mechanism of example guided artistic image generation.
Traditional style transfer methods use low-level hand-crafted features to match the patches between content image and style image~\cite{Wang:2004:EEP,Zhang:2013:STI}.
In recent years pre-trained deep convolutional neural networks are used to extract the statistical distribution of features which can capture style patterns effectively\cite{Gatys:2016:IST,Gatys:2017:CPF,Kolkin:2019:STR}.
Arbitrary style transfer methods use unified models to handle arbitrary inputs by building feed-forward architectures~\cite{Liao:2017:VAT,Li:2017:UST,huangadain,Park:2019:AST,Li:2019:LLT,deng2020arbitrary,svoboda2020two,Wu:2021:SF,Zhang:2022:EFD}.

Liu et al.~\cite{liu2021adaattn} learn spatial attention score from both shallow and deep features by an adaptive attention normalization module (AdaAttN).
An et al.~\cite{An:2021:Artflow} alleviate content leak by reversible neural flows and an unbiased feature transfer module (ArtFlow).
Chen et al.~\cite{chen2021artistic} apply internal-external scheme to learn feature statistics (mean and standard deviation) as style priors (IEST).
Zhang et al.~\cite{Zhang:2022:CAST} learn style representation directly from image features via contrastive learning to achieve domain enhanced arbitrary style transfer (CAST).
Besides CNN, visual transformer has also been used for style transfer tasks.
Wu et al.~\cite{Wu:2021:SF} perform content-guided global style composition by a transformer-driven style composition module (StyleFormer).
Deng et al.~\cite{deng2022stytr2} propose a transformer-based method  (StyTr$^2$) to avoid the biased content representation in style transfer by taking long-range dependencies of input images into account.
Image style transfer methods mainly focus on learning and transferring colors and brushstrokes, but have difficulty on other artistic creativity factors such as object shape and decorative elements.

\paragraph{Text-to-image synthesis}
Text guided synthesis methods can also be used to generate artistic images~\cite{ramesh2021zero,ding2022cogview2,imagen,dalle2,yu2022scaling}.
CLIPDraw~\cite{frans2021clipdraw} synthesizes artistic images from text by using CLIP encoder~\cite{clip} to maximize similarity between the textual description and generated drawing.
VQGAN-CLIP~\cite{crowson2022vqgan} uses CLIP-guided VQGAN~\cite{esser2021taming} to generate artistic images of various styles from text prompts.
Rombach et al.~\cite{latentdiffusion} train diffusion models~\cite{sohl2015deep,ho2020denoising} in the latent space to reduce complexity and generate high quality artistic images from texts.
Those models only use text guidance to generate an image, without fine-grained content or style control.
Some methods add image prompt to increase controllability to the content of the generate image.
CLIPstyler~\cite{CLIPstyler} transfers an input image to a desired style with a text description by using CLIP loss and PatchCLIP loss.
StyleGAN-NADA~\cite{StyleGANNADA} use CLIP to adaptively train the generator, which can transfer a photo to artistic domain by text description of the target style.
Huang et al.~\cite{Huang2022MGAD} propose a diffusion-based artistic image generation approach by utilizing multimodal prompts as guidance to control the classifier free diffusion model.
Hertz et al.~\cite{hertz2022prompt} change an image to artistic style by using the text prompt with style description and injecting the source attention maps.
Those methods are still difficult to generate images with complex or special artistic characteristics which cannot be described by normal texts.
StyleCLIPDraw~\cite{StyleCLIPDraw} jointly optimizes text description and style image for artistic image generation.
Liu et al.~\cite{liu2022name} extract style description from CLIP model by a contrastive training strategy, which enables the network to perform style transfer between content image and a textual style description.
These methods utilize the aligned image and text embedding of CLIP to achieve style transfer via narrowing the distance between the generated image and the style image, while we obtain the image representation straight from the artistic image.

\paragraph{Inversion of diffusion models}
Inversion of diffusion models is to find a noise map and a conditioning vector corresponding to a generated image.
It is a potential way for improving the quality of example guided artistic image generation.
However, naively adding noise to an image and then denoising it may yield an image with significantly different content.
Choi et al.~\cite{Choi:2021:ILVR} perform inversion by using noised low-pass filter data from the target image as the basis for the denoising process.
Dhariwal et al.~\cite{dhariwal2021diffusion} invert the deterministic DDIM~\cite{DDIM} sampling process in closed form to obtain a latent noise map that will produce a given real image.
Ramesh~\cite{dalle2} develop a text-conditional image generator based on the diffusion models and the inverted CLIP.
The above methods are difficult to generate new instances of a given example while maintaining fidelity.
Gal et al.~\cite{gal2022image} presents a textual inversion method to find a new pseudo-word to describe visual concept of a specific object or artistic style in the embedding space of a fixed text-to-image model.
They use optimization-based methods to directly optimize the embedding of the concept.
Ruiz et al.~\cite{ruiz2022dreambooth} implant a subject into the output domain of the text-to-image diffusion model so that it can be synthesized in novel views with a unique identifier.
Their inversion method is based on the fine-tuning of diffusion models, which demands high computational resources.
Both methods learn concepts from pictures through textual inversion, while they need a small (3-5) image set to depict the concept.
The concept they aim to learn is always an object.
Our method can learn the corresponding textual embedding from a single image and use it as a condition to guide the generation of artistic images without fine-tuning the generative model.

\section{Method}

\begin{figure*}
\centering
\includegraphics[width=0.99\linewidth]{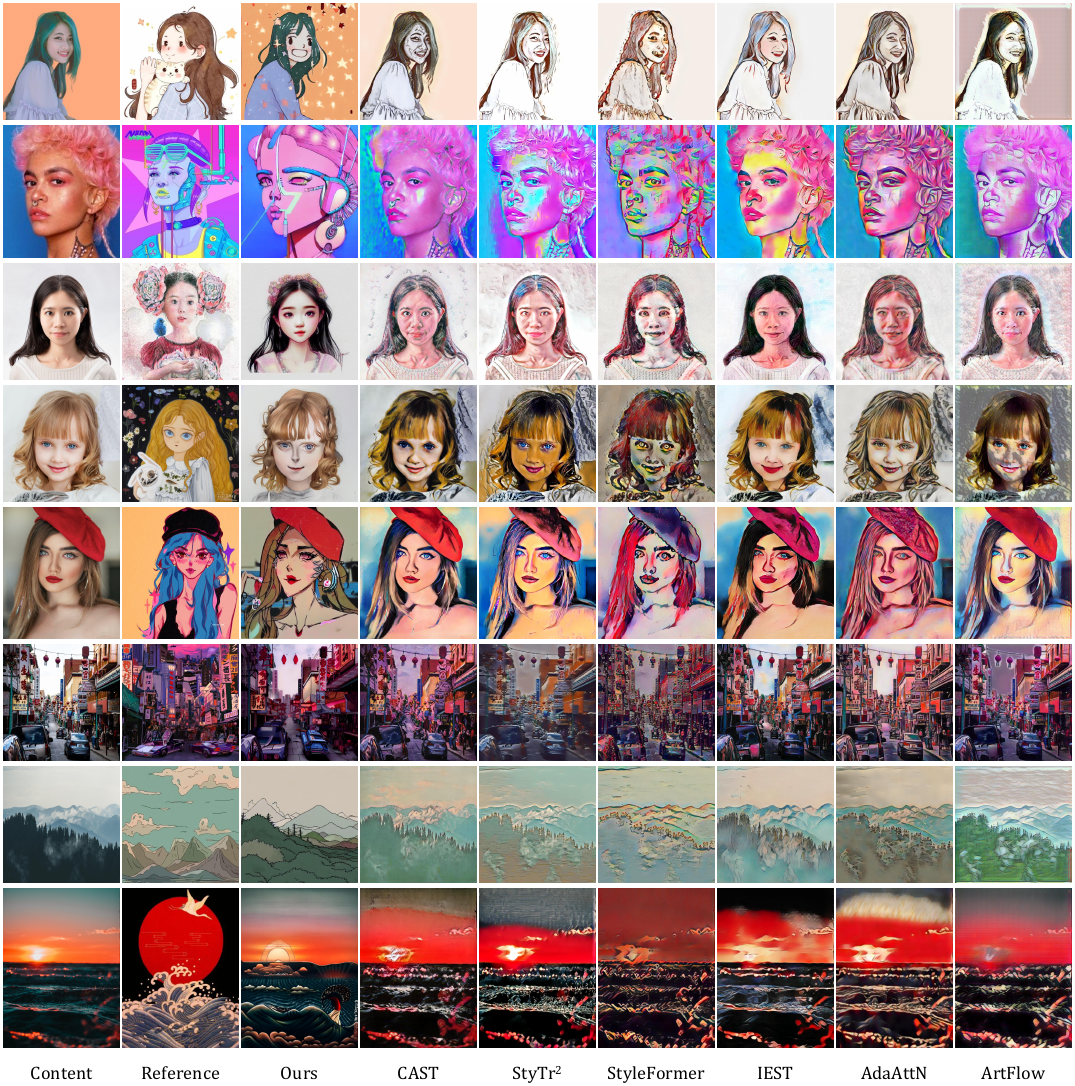}
\caption{Qualitative comparison with several state-of-the-arts image style transfer methods.
}
\label{fig:style_transfer}
\end{figure*}

\begin{figure*}
\centering
\includegraphics[width=0.99\linewidth]{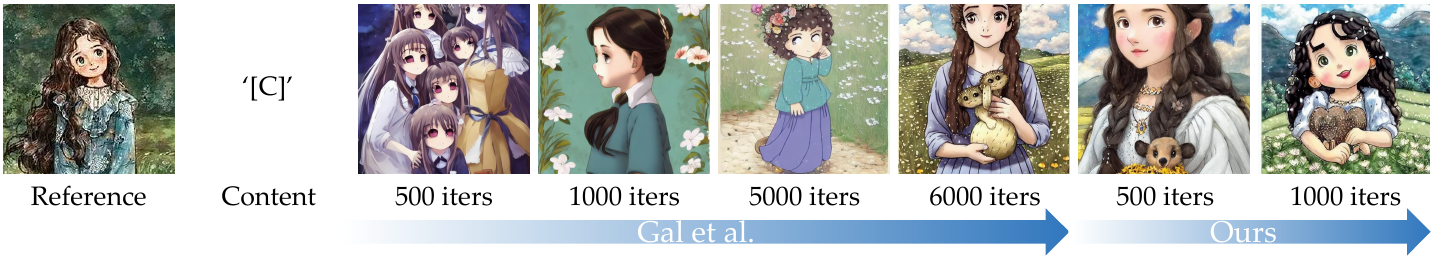}
\caption{Qualitative comparison with textual inversion~\cite{gal2022image}.
Our method outperforms in efficiency and accuracy.
It can be observed that our inversion based on the attention mechanism can converge faster to the region of text feature space corresponding to the artistic image.
}
\label{fig:textual_time}
\end{figure*}

\begin{figure*}
\centering
\vspace{-0mm}
\includegraphics[width=0.99\linewidth]{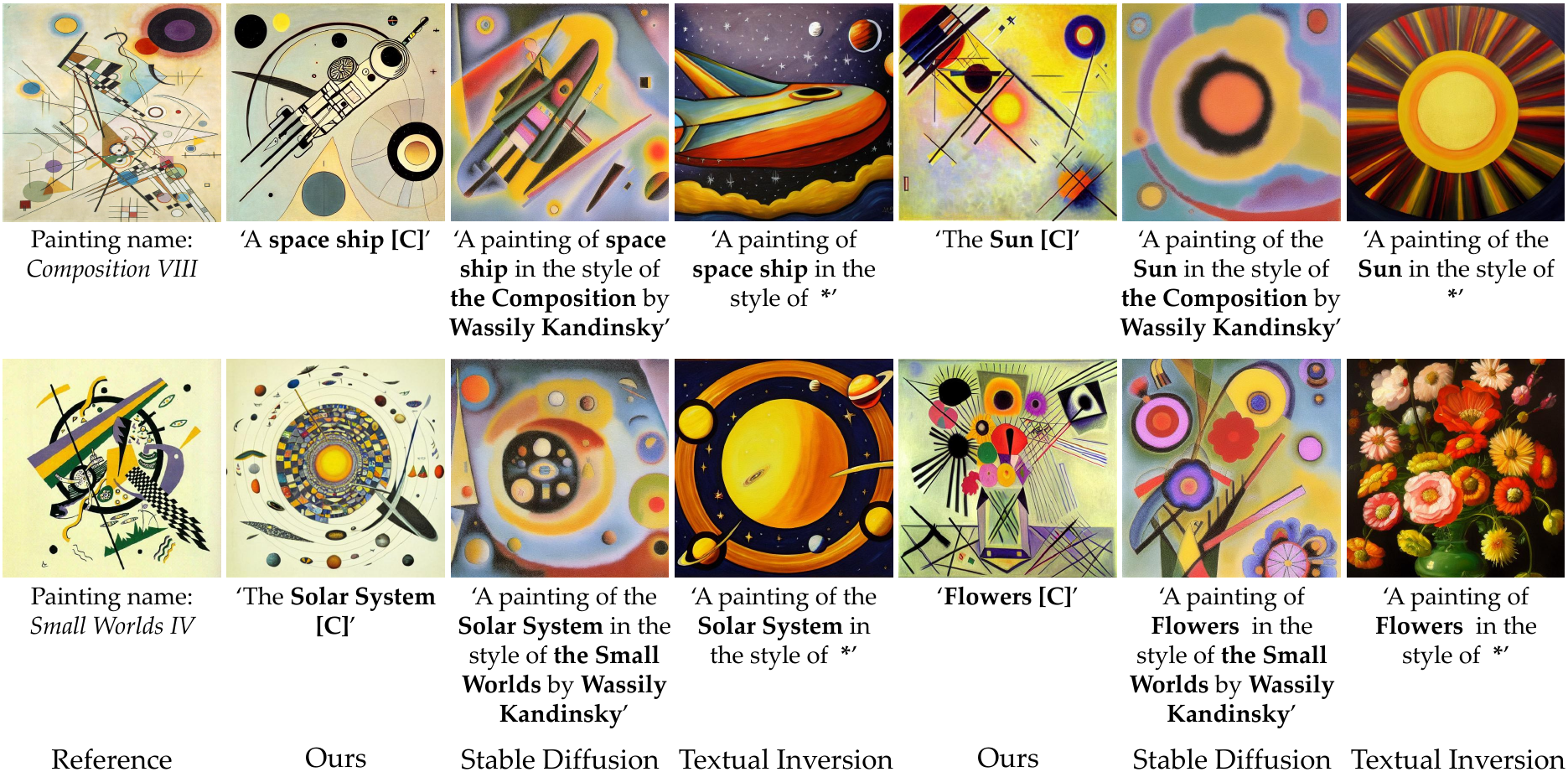}
\vspace{-2mm}
\caption{Qualitative comparison with textual inversion ~\cite{gal2022image} and SDM~\cite{latentdiffusion} in text-to-image generation.
The content of each image depends on the caption below.
``[C]'' refers to our learned description of the corresponding painting image shown in the left column.
``*'' refers to the pseudo word optimized by Textual Inversion~\cite{gal2022image}.
InST performs better in editability.
Our method is able to interact with additional text to generate artistic images with corresponding styles. 
However, when \cite{gal2022image} is trained based on a single image, it is seriously affected by the additional text and cannot maintain the style.
}
\label{fig:text2img}
\end{figure*}

\subsection{Overview}

In this work, we use inversion as the basis of our InST framework and SDM as the generative backbone.
Note that our framework is not restricted to a specific generative model.
As shown in Figure~\ref{fig:pipeline}, our method involves pixel space, latent space, and textual space.
During training, image $x$ is the same as image $y$.
The image embedding of image $x$ is obtained by the CLIP image encoder and then sent to the attention-based inversion module.
By multi-layer cross-attention, the key information of the image embedding is learned.
The inversion module gives text embedding $v$, which is converted into the standard format of caption conditioning SDMs.
Conditioned on the input textual information, the generative model obtains a series of latent codes $z_t$ through sequence denoise process from to the random noise $z_T$ and finally gives the latent code $z$ corresponding to the artistic image.
The inversion module is optimized by the simple loss of LDMs computed on the ``latent noise'' of forward process and the reverse process (see Sec.~\ref{sec:textual_inversion}).
In the inference process, $x$ is the content image, and $y$ is the reference image.
The textual embedding $v$ of the reference image $y$ guides the generative model to generate a new artistic image.

\subsection{Textual Inversion}
\label{sec:textual_inversion}

We aim to get the intermediate representation of a pre-trained text-to-image model for a specific painting.
SDMs utilize CLIP text embedding as the condition in text-to-image generation.
The CLIP text encoding contains two processes of tokenization and parameterization.
An input text is first transformed into a token, which is an index in a pre-defined dictionary, for each word or sub-word. 
After that, each token is associated with a distinct embedding vector that can be located using an index.
We set the concept of a picture as a placeholder ``[C]'', and its tokenized corresponding text embedding as a learnable vector $\hat{v}$.
$[C]$ is in the normal language domain, and $\hat{v}$ is in the textual space.
By assuming a $[C]$ that does not exist in real language, we create a ``new word'' for a certain artistic image that cannot be expressed in normal language.
To obtain $\hat{v}$, we need to design constraints as supervision that relies on a single image.
An instinctive way to learn $\hat{v}$ is by direct optimization~\cite{gal2022image}, which is minimizing the LDM loss of a single image:
\begin{equation}
\begin{aligned}
\hat{v}=\underset{v}{\arg \min } \mathbb{E}_{z,x,y,t}\left[\left\|\epsilon-\epsilon_\theta\left(z_t, t, v_\theta(y)\right)\right\|_2^2\right],
\end{aligned}
\label{eqn:ldm_loss_opt}
\end{equation}
where $y$ denotes the artistic image, $v_\theta(y)$ is a learnable vector, $z \sim E(x),\epsilon \sim \mathcal{N}(0, 1)$.
However, this optimization-based approach is inefficient, and it is difficult to obtain accurate embeddings without overfitting with a single image as training data.

Thanks to CLIP's aligned latent space of image embedding and text embedding, it provides powerful guidance for our optimization process.
We propose a learning method based on multi-layer cross attention. 
The input artistic image is first sent into the CLIP image encoder and gives image embeddings.
By performing multi-layer attention on these image embeddings, the key information of the image can be quickly obtained. 
The CLIP image encoder $\tau_\theta$ projects $y$ to an image embedding $\tau_\theta(y)$.
The multi-layer cross attention starts with $v_0 = \tau_\theta(y)$.
Then each layer is implementing $\operatorname{Attention}(Q, K, V)=\operatorname{softmax}\left(\frac{Q K^T}{\sqrt{d}}\right) \cdot V$ with:
\begin{equation}
\begin{aligned}
Q_i=W_Q^{(i)} \cdot v_i, K&=W_K^{(i)} \cdot \tau_\theta(y), V=W_V^{(i)} \cdot \tau_\theta(y),\\
v_{i+1} &= \operatorname{Attention}(Q_i, K, V).
\end{aligned}
\end{equation}
During training, the model is conditioned by the corresponding text embedding only.
To avoid overfitting, we apply a dropout strategy in each cross-attention layer which set to 0.05.

Our optimization goal can finally be defined as:
\begin{equation}
\begin{aligned}
\hat{v}=\underset{v}{\arg \min } \mathbb{E}_{z,x,y,t}\left[\left\|\epsilon-\epsilon_\theta\left(z_t, t,\operatorname{MultiAtt}(\tau_\theta(y)\right) \right\|_2^2\right],
\end{aligned}
\end{equation}
where $z \sim E(x), \epsilon \sim \mathcal{N}(0, 1)$.
$\tau_\theta$ and $\epsilon_\theta$ are fixed during training.
In this way, $\hat{v}$ can be optimized to the target area efficiently.

\subsection{Stochastic Inversion}
 
We observe that in addition to the text description, the random noise controlled by the random seed is also important for the representation of the image.
As demonstrated in \cite{hertz2022prompt}, the changes of random seed results in obvious changes of visual differences.
We define pre-trained text-to-image diffusion model-based image representation into two parts: holistic representation and detail representation.
The holistic representation refers to the text conditions, and the detail representation is controlled by the random noise.
We define the process from an image to noise maps as an inversion problem, and propose stochastic inversion to preserve the semantics of the content image.
We first add random noise to the content image, and then use the denoising U-Net in the diffusion model to predict the noise in the image.
The predicted noise is used as the initial input noise during generation to preserve content.
Specifically, for each image $z$, the stochastic inversion module takes the image latent code $z=E(y)$ as input.
Set $z_t$, the noisy version of $z$, as computable parameters, then $\epsilon_t$ is obtained by:
\begin{equation}
\begin{aligned}
\hat{\epsilon}_t&=(z_{t-1}-\mu_T(z_t,t))\sigma_t.\\
\end{aligned}
\end{equation}
We illustrate the stochastic inversion in Figure~\ref{fig:pipeline}.

\section{Experiments}
In this section, we provide visual comparisons and applications to demonstrate the effectiveness of our approach.

\paragraph{Implementation details}
We retain the original hyper-parameter choices of SDMs.
The training process takes about 20 minutes each image on one NVIDIA GeForce RTX3090 with a batch size of 1. 
The base learning rate was set to 0.001. 
The synthesis process takes the same time as SDM, which depends on the steps.

\begin{figure*}
\centering
\includegraphics[width=0.99\linewidth]{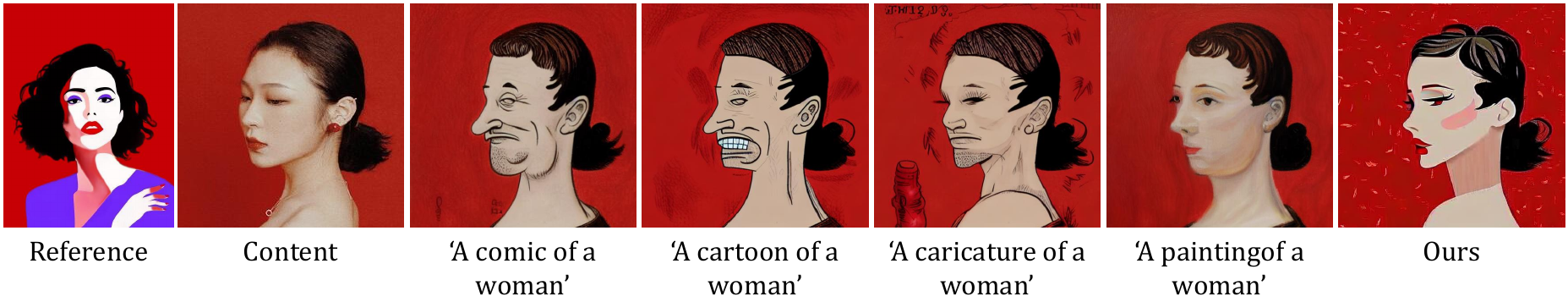}
\caption{Qualitative comparison with SDM~\cite{latentdiffusion} conditioned on human captions.
Our method can accurately represent the target style image, while describing artistic attributes of a specific painting in words for SDM is more difficult.
}
\label{fig:compare_sd}
\end{figure*}

\subsection{Comparison with Style Transfer Methods}
We compare our method with the state-of-the-arts image style transfer methods, including ArtFlow~\cite{An:2021:Artflow}, AdaAttN~\cite{liu2021adaattn}, StyleFormer~\cite{Wu:2021:SF}, IEST~\cite{chen2021artistic}, StyTr$^2$~\cite{deng2022stytr2} and CAST~\cite{Zhang:2022:CAST} to show the effectiveness of our method.
From the results, we can see apparent advantages of our method on transferring the semantics and artistic techniques of the reference images to the content images over traditional style transfer methods.
For example, our method can better transfer the shapes of important objects, such as the facial forms and eyes (the 1st, 3rd, 4th and 5th rows), the mountain (the 7th row) and the sun (the 8th row).
Our method can capture some special semantics of the reference images and reproduce the visual effects in the results, such as the stars on the background (the 1st row), the flower headwear (the 3rd row) and the roadsters (the 6th row, the cars in the content image are changed to roadsters).
Those effects are very difficult for traditional style transfer methods to achieve.

\subsection{Comparison with Text-Guided Methods}

We compare our method with textual inversion~\cite{gal2022image} and SDM~\cite{latentdiffusion} guided by a human caption.
Following ~\cite{gal2022image}, we measure \textit{accuracy} and \textit{editability} by the similarities between the CLIP embeddings of style images and generated images, and the similarities of guide texts and generated images, respectively.

\begin{table}
\centering
\caption{CLIP-based evaluations.}
\vskip -4mm
\resizebox{0.47\textwidth}{!}{
\begin{tabular}{c|ccccc}
\toprule
    & Ours & ~\cite{gal2022image} & ~\cite{latentdiffusion} & MLP & w/o drop \\
\hline
Accuracy(\%)$\uparrow$ & \textbf{78.92} & 68.84 & 63.83 & 71.03 & 72.25\\
\hline
Editability(\%)$\uparrow$ & \textbf{80.72} & 69.03 & - & 66.99 & 66.53  \\
\bottomrule
\end{tabular}}
\label{tab:clip_evaluation}
\end{table}

\paragraph{Comparison with Textual Inversion}

We begin by demonstrating the effectiveness of our attention-based inversion on learning and transferring style.
In Figure~\ref{fig:textual_time}, we show the optimization process of ~\cite{gal2022image} and ours.
Our method can quickly optimize to the target text embedding in about 1000 iterations, while \cite{gal2022image} usually takes 10 times the iterations of ours due to its simple optimization-based scheme.
In Figure~\ref{fig:text2img}, we demonstrate our superior generality and editability by giving additional semantic descriptions that do not appear in the reference image.
It can be easily observed that our method is more robust to those additional descriptions and is able to generate results that match both the textual description and the reference image.
However, \cite{gal2022image} loses adaptability to these texts and is not able to depict the specific artistic visual effect.
As shown in Table.~\ref{tab:clip_evaluation}, our method outperforms ~\cite{gal2022image} in both \textit{accuracy} and \textit{editability}.

\begin{figure}
\centering
\includegraphics[width=0.99\linewidth]{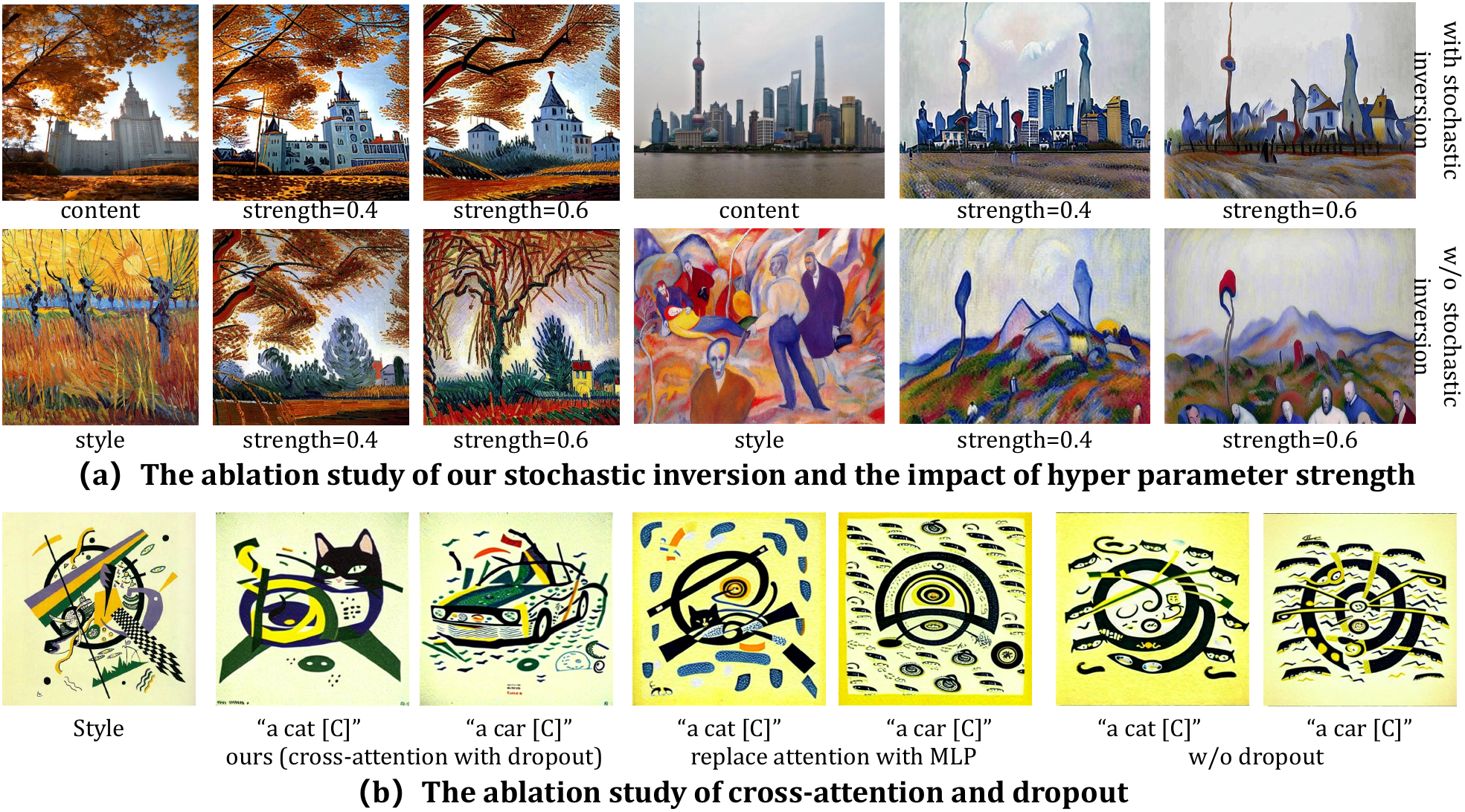}
\vspace{-3mm}
\caption{
(a) The results of the ablation study on our stochastic inversion, hyper-parameter \textit{Strength}, (b) attention module and dropout.
}
\vskip -4mm
\label{fig:ablation_camera}
\end{figure}

\begin{table*}[h]
\centering
\caption{Quantitative evaluation.
The results show the average percentage of cases in which the result of the corresponding method is preferred compared with ours.
The best results are in \textbf{bold}.
}
\resizebox{0.98\linewidth}{!}{%
\begin{tabular}{c||cccccc|c|c}
\toprule
 & CAST~\cite{Zhang:2022:CAST} & StyTr$^2$~\cite{deng2022stytr2} & StyleFormer~\cite{Wu:2021:SF} & IEST~\cite{chen2021artistic} & AdaAttN~\cite{liu2021adaattn} & ArtFlow~\cite{An:2021:Artflow}&\multicolumn{2}{c}{Textual Inversion~\cite{gal2022image}} \\ \cline{8-9}
 &  &  &  &  &  &  & img2img & txt2img \\ \hline\hline
\textbf{Preference$\uparrow$} & 0.368 & 0.310 & 0.218 & 0.161 & 0.310 & 0.276 & 0.379 & 0.121 \\ \hline
\textbf{Ours} & \textbf{0.632} & \textbf{0.690} & \textbf{0.782} & \textbf{0.839} & \textbf{0.690 }& \textbf{0.724} & \textbf{0.621} & \textbf{0.879}\\
\bottomrule
\end{tabular}
}
\label{tab:user}
\end{table*}

\paragraph{Comparison with SDMs}
We compare with the state-of-the-art text-to-image generative model SDM~\cite{latentdiffusion}.
SDM can generate high-quality images from text descriptions. However, it is difficult to describe the style of a specific painting with only text as a condition, so satisfactory results cannot be obtained.
As shown in Figure~\ref{fig:text2img} and Figure~\ref{fig:compare_sd}, our method better captures the unique artistic attributes of the reference images. 
As shown in Table~\ref{tab:clip_evaluation}, our method outperforms ~\cite{latentdiffusion} in \textit{accuracy}.

\subsection{Ablation Study}

\paragraph{Stochastic inversion}
As shown in Figure~\ref{fig:ablation_camera}(a), the buildings in the content images are turned into trees or mountains without stochastic inversion, while the full model can maintain the content information and reduce the impact of the semantic of the style image.

\paragraph{Hyper-parameter \textit{Strength}}
For image synthesis, the most related hyper-parameter is the \textit{strength} of changes and its impacts are shown in Figure~\ref{fig:ablation_camera}(a).
The larger the \textit{Strength}, the stronger the influence of the style image on the generated result, vice versa, the generated image is closer to the content image

\paragraph{Attention module}
We show the ablation study of multi-layer attention  Figures~\ref{fig:ablation_camera}(b).
By interacting with CLIP embedding multiple times, the multi-layer attention helps the learned concept be consistent with the CLIP feature space, such improve editability, as shown in Table.~\ref{tab:clip_evaluation}.

\paragraph{Dropout}
Dropout is added in the linear layer of attention module to prevent overfitting.
As shown in Figure~\ref{fig:ablation_camera}(b) and Table~\ref{tab:clip_evaluation}, by dropping the parameters of the latent embeddings, both the accuracy and the editability are improved.

\subsection{User Study}
\label{sec:user_study}

We compare our method with several SOTA image style transfer methods (i.e., ArtFlow~\cite{An:2021:Artflow}, AdaAttN~\cite{liu2021adaattn}, StyleFormer~\cite{Wu:2021:SF}, IEST~\cite{chen2021artistic}, StyTr$^2$~\cite{deng2022stytr2}, and CAST~\cite{Zhang:2022:CAST}), and text-to-image generation method (i.e., Textual Inversion~\cite{gal2022image}).
All the baselines are trained using publicly available implementations with default configurations.

For each participant, 26 content-reference pairs are randomly selected and the generated results of ours and one of the other methods are displayed randomly.
Participants were suggested that the artistic consistency between the generated image and the reference image was the main metric.
Then, they were invited to select the better result of each content-reference pair.
Finally, we collect 2,262 votes from 87 participants.
The percentage of votes for each approach is shown in Table~\ref{tab:user}, demonstrating that our method achieves the best visual characteristics transfer results.

Furthermore, we conducted a survey of 60 participants on the preferences of the content image guidance strength and artistic visual effects.
In the case of a content image existing, users tend to consider that ``To depict the artistic style, the details of the content should be embellished appropriately''.
We then invite the participants to rank the factors of their expected visual effect.
The average comprehensive score of the options in the sorting question is automatically calculated based on the ranking of the options by all the participants. 
The higher the score, the higher the comprehensive ranking. 
The scoring rule is:
\begin{equation}
\begin{aligned}
score = \frac{(\sum frequency \times weight) }{participantes},
\end{aligned}
\end{equation}
where $score$ denotes the average comprehensive score of the options, $participantes$ denotes the number of people who complete this question, $frequency$ denotes the frequency that the option is selected by users, $weight$ denotes the weight which is determined by the option's ranking.
The ranking results (rank by score from highest to lowest):
(1) Similar artistic effect on semantic corresponding subjects (s=5.4);
(2) With the same paint material ($score$=3.65);
(3) Having similar brushstrokes ($score$=3.2);
(4) Having typical shapes ($score$=2.65);
(5) With the same decorative elements ($score$=2.1);
(6) Sharing the same color ($score$=1.4).

\subsection{Discussions and Limitations}

\begin{figure}
\centering
\includegraphics[width=0.99\linewidth]{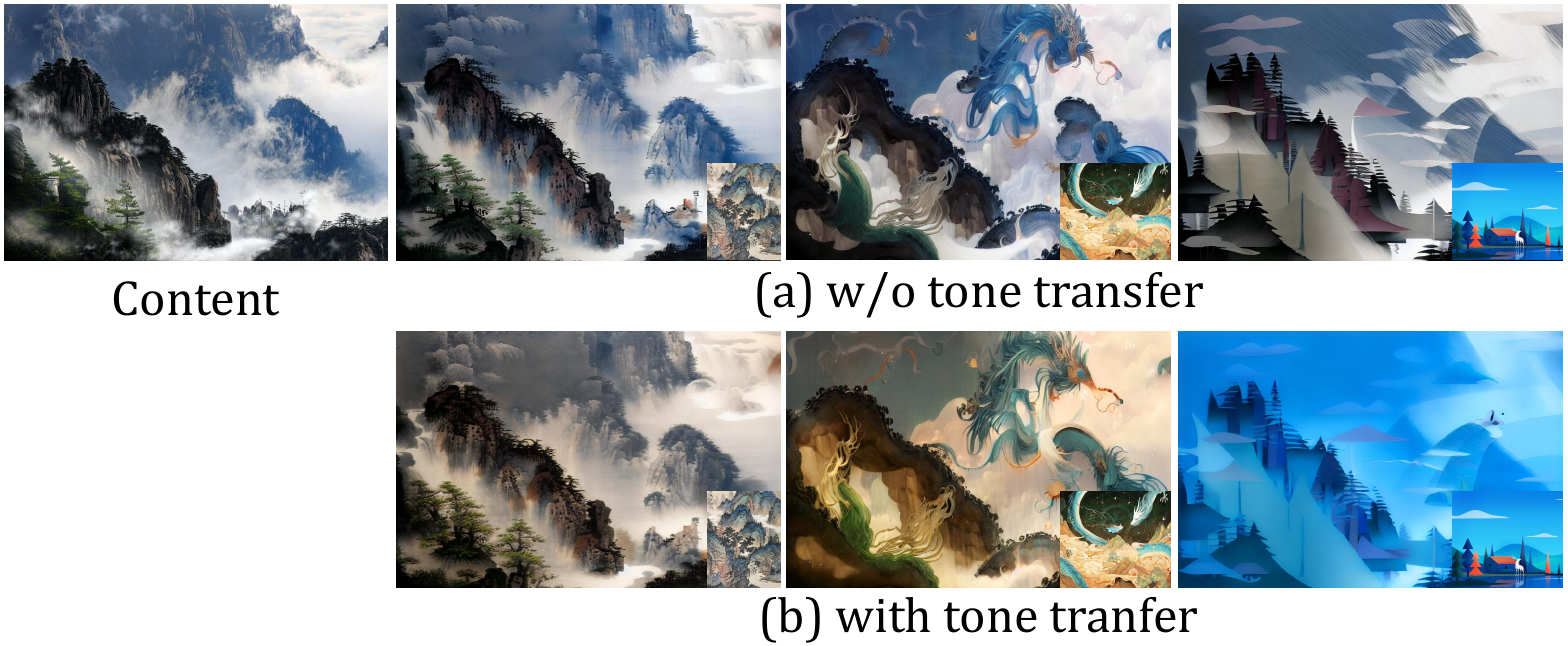}
\caption{For different art forms, the tone of natural images and of artistic images have their own advantages.}
\vspace{-3mm}
\label{fig:color}
\end{figure}

Although our method can transfer typical colors to some extent, when there is a significant difference between the colors of the content image and the reference image, our method may fail to transfer the color in a one-to-one correspondence semantically. For example, the green hair of the content images in the 1st row of Figure~\ref{fig:style_transfer} is not transferred into brown.
As shown in Figure~\ref{fig:color}, we employ an additional tone transfer module~\cite{huangadain} to align the color of content and reference images.
However, we observe that different users have different preferences on whether the colors of the content image should be retained.
We believe that the colors of a photograph is crucial, so we choose to respect the tone of the original content image in some conditions.

\section{Conclusion}

We introduce a novel example-guided artistic image generation framework called InST, which refers to learning the high-level textual descriptions of a single painting image and then guiding the text-to-image generative model in creating images of specific artistic appearance. 
We propose an attention-based textual inversion method to invert a painting into the corresponding textual embeddings, which benefits from the aligned text and image feature spaces of CLIP. 
The extensive experimental results demonstrate that our method achieves superior image-to-image and text-to-image generation results compared with state-of-the-art approaches. 
Our approach is intended to pave the way for upcoming unique artistic image synthesis tasks.

\small{
\paragraph{Acknowledgment}
This work was supported in part by National Key R\&D Program of China under no. 2020AAA0106200, by National Natural Science Foundation of China under nos. 61832016, U20B2070, and 62102162, and in part by Beijing Natural Science Foundation under no. L221013.
}

{\small
\bibliographystyle{ieee_fullname}
\bibliography{CreativityTransfer}

\begin{thebibliography}{10}\itemsep=-1pt

\bibitem{An:2021:Artflow}
Jie An, Siyu Huang, Yibing Song, Dejing Dou, Wei Liu, and Jiebo Luo.
\newblock {ArtFlow}: Unbiased image style transfer via reversible neural flows.
\newblock In {\em IEEE/CVF Conferences on Computer Vision and Pattern
  Recognition (CVPR)}, pages 862--871, 2021.

\bibitem{chefer2022image}
Hila Chefer, Sagie Benaim, Roni Paiss, and Lior Wolf.
\newblock Image-based clip-guided essence transfer.
\newblock In {\em European Conference on Computer Vision (ECCV)}, pages
  695--711. Springer, 2022.

\bibitem{chen2021artistic}
Haibo Chen, Lei Zhao, Zhizhong Wang, Zhang~Hui Ming, Zhiwen Zuo, Ailin Li, Wei
  Xing, and Dongming Lu.
\newblock Artistic style transfer with internal-external learning and
  contrastive learning.
\newblock In {\em Advances in Neural Information Processing Systems (NeurIPS)},
  2021.

\bibitem{Choi:2021:ILVR}
Jooyoung Choi, Sungwon Kim, Yonghyun Jeong, Youngjune Gwon, and Sungroh Yoon.
\newblock Ilvr: Conditioning method for denoising diffusion probabilistic
  models.
\newblock In {\em IEEE/CVF International Conference on Computer Vision (ICCV)},
  pages 14347--14356, 2021.

\bibitem{crowson2022vqgan}
Katherine Crowson, Stella Biderman, Daniel Kornis, Dashiell Stander, Eric
  Hallahan, Louis Castricato, and Edward Raff.
\newblock {VQGAN-CLIP}: Open domain image generation and editing with natural
  language guidance.
\newblock In {\em European Conference on Computer Vision (ECCV)}, pages
  88--105. Springer, 2022.

\bibitem{deng2022stytr2}
Yingying Deng, Fan Tang, Weiming Dong, Chongyang Ma, Xingjia Pan, Lei Wang, and
  Changsheng Xu.
\newblock {StyTr}$^2$: Image style transfer with transformers.
\newblock In {\em IEEE/CVF Conference on Computer Vision and Pattern
  Recognition (CVPR)}, pages 11326--11336, 2022.

\bibitem{deng2020arbitrary}
Yingying Deng, Fan Tang, Weiming Dong, Wen Sun, Feiyue Huang, and Changsheng
  Xu.
\newblock Arbitrary style transfer via multi-adaptation network.
\newblock In {\em ACM International Conference on Multimedia}, pages
  2719--2727, 2020.

\bibitem{Guided-Diffusion}
Prafulla Dhariwal and Alex Nichol.
\newblock Diffusion models beat {GANs} on image synthesis.
\newblock In {\em Advances Neural Information Processing Systems (NeurIPS)},
  pages 8780--8794, 2021.

\bibitem{dhariwal2021diffusion}
Prafulla Dhariwal and Alexander Nichol.
\newblock Diffusion models beat {GANs} on image synthesis.
\newblock {\em Advances in Neural Information Processing Systems (NeurIPS)},
  34:8780--8794, 2021.

\bibitem{ding2022cogview2}
Ming Ding, Wendi Zheng, Wenyi Hong, and Jie Tang.
\newblock {CogView2}: Faster and better text-to-image generation via
  hierarchical transformers.
\newblock {\em arXiv preprint arXiv:2204.14217}, 2022.

\bibitem{esser2021taming}
Patrick Esser, Robin Rombach, and Bjorn Ommer.
\newblock Taming transformers for high-resolution image synthesis.
\newblock In {\em IEEE/CVF Conference on Computer Vision and Pattern
  Recognition (CVPR)}, pages 12873--12883, 2021.

\bibitem{frans2021clipdraw}
Kevin Frans, Lisa~B Soros, and Olaf Witkowski.
\newblock {CLIPDraw}: Exploring text-to-drawing synthesis through
  language-image encoders.
\newblock {\em arXiv preprint arXiv:2106.14843}, 2021.

\bibitem{fu2022ldast}
Tsu-Jui Fu, Xin~Eric Wang, and William~Yang Wang.
\newblock Language-driven artistic style transfer.
\newblock In {\em European Conference on Computer Vision (ECCV)}, pages
  717--734, 2022.

\bibitem{gal2022image}
Rinon Gal, Yuval Alaluf, Yuval Atzmon, Or Patashnik, Amit~H Bermano, Gal
  Chechik, and Daniel Cohen-Or.
\newblock An image is worth one word: Personalizing text-to-image generation
  using textual inversion.
\newblock {\em arXiv preprint arXiv:2208.01618}, 2022.

\bibitem{StyleGANNADA}
Rinon Gal, Or Patashnik, Haggai Maron, Amit~H. Bermano, Gal Chechik, and Daniel
  Cohen-Or.
\newblock {StyleGAN-NADA: CLIP}-guided domain adaptation of image generators.
\newblock {\em ACM Transactions on Graphics}, 41(4):141:1--141:13, 2022.

\bibitem{gatyscnnstyle}
Leon~A Gatys, Alexander~S Ecker, and Matthias Bethge.
\newblock Image style transfer using convolutional neural networks.
\newblock In {\em IEEE/CVF Conferences on Computer Vision and Pattern
  Recognition (CVPR)}, pages 2414--2423, 2016.

\bibitem{Gatys:2016:IST}
Leon~A Gatys, Alexander~S Ecker, and Matthias Bethge.
\newblock Image style transfer using convolutional neural networks.
\newblock In {\em IEEE Conference on Computer Vision and Pattern Recognition
  (CVPR)}, pages 2414--2423, 2016.

\bibitem{Gatys:2017:CPF}
Leon~A. Gatys, Alexander~S. Ecker, Matthias Bethge, Aaron Hertzmann, and Eli
  Shechtman.
\newblock Controlling perceptual factors in neural style transfer.
\newblock In {\em IEEE Conference on Computer Vision and Pattern Recognition
  (CVPR)}, pages 3730--3738, 2017.

\bibitem{hertz2022prompt}
Amir Hertz, Ron Mokady, Jay Tenenbaum, Kfir Aberman, Yael Pritch, and Daniel
  Cohen-Or.
\newblock Prompt-to-prompt image editing with cross attention control.
\newblock {\em arXiv preprint arXiv:2208.01626}, 2022.

\bibitem{DDPM}
Jonathan Ho, Ajay Jain, and Pieter Abbeel.
\newblock Denoising diffusion probabilistic models.
\newblock {\em Advances in Neural Information Processing Systems (NeurIPS)},
  pages 6840--6851, 2020.

\bibitem{ho2020denoising}
Jonathan Ho, Ajay Jain, and Pieter Abbeel.
\newblock Denoising diffusion probabilistic models.
\newblock {\em Advances in Neural Information Processing Systems (NeurIPS)},
  33:6840--6851, 2020.

\bibitem{Huang2022MGAD}
Nisha Huang, Fan Tang, Weiming Dong, and Changsheng Xu.
\newblock Draw your art dream: Diverse digital art synthesis with multimodal
  guided diffusion.
\newblock In {\em ACM International Conference on Multimedia}, page
  1085–1094, 2022.

\bibitem{huangadain}
Xun Huang and Serge Belongie.
\newblock Arbitrary style transfer in real-time with adaptive instance
  normalization.
\newblock In {\em IEEE International Conference on Computer Vision (ICCV)},
  pages 1501--1510, 2017.

\bibitem{imagic}
Bahjat Kawar, Shiran Zada, Oran Lang, Omer Tov, Huiwen Chang, Tali Dekel, Inbar
  Mosseri, and Michal Irani.
\newblock Imagic: Text-based real image editing with diffusion models.
\newblock {\em arXiv preprint arXiv:2210.09276}, 2022.

\bibitem{Kolkin:2019:STR}
Nicholas Kolkin, Jason Salavon, and Gregory Shakhnarovich.
\newblock Style transfer by relaxed optimal transport and self-similarity.
\newblock In {\em IEEE/CVF Conference on Computer Vision and Pattern
  Recognition (CVPR)}, pages 10043--10052, 2019.

\bibitem{CLIPstyler}
Gihyun Kwon and Jong~Chul Ye.
\newblock {CLIPstyler}: Image style transfer with a single text condition.
\newblock In {\em IEEE/CVF Conference on Computer Vision and Pattern
  Recognition (CVPR)}, page 18062–18071, 2022.

\bibitem{Li:2019:LLT}
Xueting Li, Sifei Liu, Jan Kautz, and Ming-Hsuan Yang.
\newblock Learning linear transformations for fast image and video style
  transfer.
\newblock In {\em IEEE/CVF Conference on Computer Vision and Pattern
  Recognition (CVPR)}, pages 3804--3812, 2019.

\bibitem{Liao:2017:VAT}
Jing Liao, Yuan Yao, Lu Yuan, Gang Hua, and Sing~Bing Kang.
\newblock Visual attribute transfer through deep image analogy.
\newblock {\em ACM Transactions on Graphics}, 36(4), 2017.

\bibitem{Li:2017:UST}
Yijun Li, Chen Fang, Jimei Yang, Zhaowen Wang, Xin Lu, and Ming-Hsuan Yang.
\newblock Universal style transfer via feature transforms.
\newblock In {\em Advances Neural Information Processing Systems (NeurIPS)},
  pages 386--396, 2017.

\bibitem{liu2021adaattn}
Songhua Liu, Tianwei Lin, Dongliang He, Fu Li, Meiling Wang, Xin Li, Zhengxing
  Sun, Qian Li, and Errui Ding.
\newblock {AdaAttN}: Revisit attention mechanism in arbitrary neural style
  transfer.
\newblock In {\em IEEE/CVF International Conference on Computer Vision (ICCV)},
  pages 6649--6658, 2021.

\bibitem{liu2022name}
Zhi-Song Liu, Li-Wen Wang, Wan-Chi Siu, and Vicky Kalogeiton.
\newblock Name your style: An arbitrary artist-aware image style transfer.
\newblock {\em arXiv preprint arXiv:2202.13562}, 2022.

\bibitem{nichol2022glide}
Alex Nichol, Prafulla Dhariwal, Aditya Ramesh, Pranav Shyam, Pamela Mishkin,
  Bob McGrew, Ilya Sutskever, and Mark Chen.
\newblock {GLIDE}: Towards photorealistic image generation and editing with
  text-guided diffusion models.
\newblock In {\em International Conference on Machine Learning (ICML)}, 2022.

\bibitem{Park:2019:AST}
Dae~Young Park and Kwang~Hee Lee.
\newblock Arbitrary style transfer with style-attentional networks.
\newblock In {\em IEEE/CVF Conference on Computer Vision and Pattern
  Recognition (CVPR)}, pages 5880--5888, 2019.

\bibitem{StyleCLIP}
Or Patashnik, Zongze Wu, Eli Shechtman, Daniel Cohen-Or, and Dani Lischinski.
\newblock {StyleCLIP}: Text-driven manipulation of stylegan imagery.
\newblock In {\em IEEE/CVF International Conference on Computer Vision (ICCV)},
  pages 2085--2094, 2021.

\bibitem{clip}
Alec Radford, Jong~Wook Kim, Chris Hallacy, Aditya Ramesh, Gabriel Goh,
  Sandhini Agarwal, Girish Sastry, Amanda Askell, Pamela Mishkin, Jack Clark,
  et~al.
\newblock Learning transferable visual models from natural language
  supervision.
\newblock In {\em International Conference on Machine Learning (ICML)}, pages
  8748--8763, 2021.

\bibitem{dalle2}
Aditya Ramesh, Prafulla Dhariwal, Alex Nichol, Casey Chu, and Mark Chen.
\newblock Hierarchical text-conditional image generation with clip latents.
\newblock {\em arXiv preprint arXiv:2204.06125}, 2022.

\bibitem{ramesh2021zero}
Aditya Ramesh, Mikhail Pavlov, Gabriel Goh, Scott Gray, Chelsea Voss, Alec
  Radford, Mark Chen, and Ilya Sutskever.
\newblock Zero-shot text-to-image generation.
\newblock In {\em International Conference on Machine Learning}, pages
  8821--8831. PMLR, 2021.

\bibitem{latentdiffusion}
Robin Rombach, Andreas Blattmann, Dominik Lorenz, Patrick Esser, and Bj{\"o}rn
  Ommer.
\newblock High-resolution image synthesis with latent diffusion models.
\newblock In {\em IEEE/CVF Conference on Computer Vision and Pattern
  Recognition (CVPR)}, pages 10684--10695, 2022.

\bibitem{ruiz2022dreambooth}
Nataniel Ruiz, Yuanzhen Li, Varun Jampani, Yael Pritch, Michael Rubinstein, and
  Kfir Aberman.
\newblock {DreamBooth}: Fine tuning text-to-image diffusion models for
  subject-driven generation.
\newblock {\em arXiv preprint arXiv:2208.12242}, 2022.

\bibitem{imagen}
Chitwan Saharia, William Chan, Saurabh Saxena, Lala Li, Jay Whang, Emily
  Denton, Seyed Kamyar~Seyed Ghasemipour, Burcu~Karagol Ayan, S~Sara Mahdavi,
  Rapha~Gontijo Lopes, et~al.
\newblock Photorealistic text-to-image diffusion models with deep language
  understanding.
\newblock {\em arXiv preprint arXiv:2205.11487}, 2022.

\bibitem{StyleCLIPDraw}
Peter Schaldenbrand, Zhixuan Liu, and Jean Oh.
\newblock {StyleCLIPDraw}: Coupling content and style in text-to-drawing
  translation.
\newblock In {\em International Joint Conference on Artificial Intelligence
  (IJCAI)}, pages 4966--4972. International Joint Conferences on Artificial
  Intelligence Organization, 7 2022.

\bibitem{sohl2015deep}
Jascha Sohl-Dickstein, Eric Weiss, Niru Maheswaranathan, and Surya Ganguli.
\newblock Deep unsupervised learning using nonequilibrium thermodynamics.
\newblock In {\em International Conference on Machine Learning (ICML)}, pages
  2256--2265. PMLR, 2015.

\bibitem{DDIM}
Jiaming Song, Chenlin Meng, and Stefano Ermon.
\newblock Denoising diffusion implicit models.
\newblock In {\em International Conference on Learning Representations (ICLR)},
  2021.

\bibitem{svoboda2020two}
Jan Svoboda, Asha Anoosheh, Christian Osendorfer, and Jonathan Masci.
\newblock Two-stage peer-regularized feature recombination for arbitrary image
  style transfer.
\newblock In {\em IEEE/CVF Conference on Computer Vision and Pattern
  Recognition}, pages 13816--13825, 2020.

\bibitem{von-platen-etal-2022-diffusers}
Patrick von Platen, Suraj Patil, Anton Lozhkov, Pedro Cuenca, Nathan Lambert,
  Kashif Rasul, Mishig Davaadorj, and Thomas Wolf.
\newblock Diffusers: State-of-the-art diffusion models.
\newblock https://github.com/huggingface/diffusers, 2022.

\bibitem{Wang:2004:EEP}
Bin Wang, Wenping Wang, Huaiping Yang, and Jiaguang Sun.
\newblock Efficient example-based painting and synthesis of {2D} directional
  texture.
\newblock {\em IEEE Transactions on Visualization and Computer Graphics},
  10(3):266--277, 2004.

\bibitem{wu2022creative}
Xianchao Wu.
\newblock Creative painting with latent diffusion models.
\newblock {\em arXiv preprint arXiv:2209.14697}, 2022.

\bibitem{Wu:2021:SF}
Xiaolei Wu, Zhihao Hu, Lu Sheng, and Dong Xu.
\newblock {StyleFormer}: Real-time arbitrary style transfer via parametric
  style composition.
\newblock In {\em IEEE/CVF International Conference on Computer Vision (ICCV)},
  pages 14598--14607, 2021.

\bibitem{yu2022scaling}
Jiahui Yu, Yuanzhong Xu, Jing~Yu Koh, Thang Luong, Gunjan Baid, Zirui Wang,
  Vijay Vasudevan, Alexander Ku, Yinfei Yang, Burcu~Karagol Ayan, et~al.
\newblock Scaling autoregressive models for content-rich text-to-image
  generation.
\newblock {\em arXiv preprint arXiv:2206.10789}, 2022.

\bibitem{Zhang:2013:STI}
Wei Zhang, Chen Cao, Shifeng Chen, Jianzhuang Liu, and Xiaoou Tang.
\newblock Style transfer via image component analysis.
\newblock {\em IEEE Transactions on Multimedia}, 15(7):1594--1601, 2013.

\bibitem{Zhang:2022:EFD}
Yabin Zhang, Minghan Li, Ruihuang Li, Kui Jia, and Lei Zhang.
\newblock Exact feature distribution matching for arbitrary style transfer and
  domain generalization.
\newblock In {\em IEEE/CVF Conference on Computer Vision and Pattern
  Recognition (CVPR)}, pages 8035--8045, June 2022.

\bibitem{Zhang:2022:CAST}
Yuxin Zhang, Fan Tang, Weiming Dong, Haibin Huang, Chongyang Ma, Tong-Yee Lee,
  and Changsheng Xu.
\newblock Domain enhanced arbitrary image style transfer via contrastive
  learning.
\newblock In {\em ACM SIGGRAPH 2022 Conference Proceedings}, pages 12:1--12:8,
  New York, NY, USA, 2022. Association for Computing Machinery.

\end{thebibliography}
}

\end{document}